# Circle detection by Harmony Search Optimization

Erik Cuevas, Noé Ortega-Sánchez, Daniel Zaldivar[1] and Marco Pérez-Cisneros

Departamento de Ciencias Computacionales
Universidad de Guadalajara, CUCEI
Av. Revolución 1500, Guadalajara, Jal, México
{erik.cuevas, [1]daniel.zaldivar, marco.perez}cucei.udg.mx

**Abstract**

Automatic circle detection in digital images has received considerable attention over the last years in computer vision as several efforts have aimed for an optimal circle detector. This paper presents an algorithm for automatic detection of circular shapes that considers the overall process as an optimization problem. The approach is based on the Harmony Search Algorithm (HSA), a derivative free meta-heuristic optimization algorithm inspired by musicians while improvising new harmonies. The algorithm uses the encoding of three points as candidate circles (harmonies) over the edge-only image. An objective function evaluates (harmony quality) if such candidate circles are actually present in the edge image. Guided by the values of this objective function, the set of encoded candidate circles are evolved using the HSA so that they can fit to the actual circles on the edge map of the image (optimal harmony). Experimental results from several tests on synthetic and natural images with a varying complexity range have been included to validate the efficiency of the proposed technique regarding accuracy, speed and robustness.

*Keywords*: Circle detection; Harmony Search Algorithm; Meta-heuristic Algorithms; intelligent image processing.

## 1. Introduction

The problem of detecting circular features arises inside many areas of image analysis, being particularly relevant for some industrial applications such as automatic inspection of manufactured products and components, aided vectorization of drawings and target detection among others [1]. Two sorts of techniques are commonly applied to solve the object location challenge: first hand deterministic techniques including the application of Hough transform based methods [2], geometric hashing and template or model matching techniques [3,4]. On the other hand, stochastic techniques including random sample consensus techniques [5], simulated annealing [6] and Genetic Algorithms (GA) [7] have been also used.

Template and model matching techniques are the first approaches to be applied to shape detection yielding a considerable amount of publications [8]. Shape coding techniques and combination of shape properties have been commonly used to represent objects. Their main drawback is related to the contour extraction step from a real image which hardly deals with pose invariance, except for very simple objects.

Circle detection in digital images is commonly performed by the Circular Hough Transform [9]. A typical Hough-based approach employs edge information obtained by means of an edge detector to infer locations and radius values. Peak detection is then performed by averaging, filtering and histo-gramming the transformed space. However, such an approach requires a large storage space given the required 3-D cells to cover all parameters ($x, y, r$) and a high computational complexity which yields a low processing speed. The accuracy of the extracted parameters for the detected circle is poor, particularly under the presence of noise [3]. For a digital image holding a significant width and height and a densely populated edge pixel map, the required processing time for Circular Hough Transform makes it prohibitive to be deployed in real time applications. In order to overcome this problem, some other researchers have proposed new approaches based on the Hough transform, for instance the probabilistic Hough transform [11], the randomized Hough transform (RHT) [12] and the fuzzy Hough transform [13]. Alternative transformations have also been presented in literature as the one proposed by Becker in [6]. Although those new approaches demonstrated faster processing speeds in comparison to the original Hough Transform, they are still highly sensitive to noise.

---

[1] Corresponding author, Tel +52 33 1378 5900, ext. 7715, E-mail: daniel.zaldivar@cucei.udg.mx

Researchers have started recently to investigate evolutionary algorithms as an alternative way to perform circle detection. Such approaches involve the use of Genetic Algorithms (GA) [15], the Bacterial Foraging Algorithm (BFAO) [16] or the Differential Evolution technique (DE) [17]. For such methods, few assumptions are made about the objective function or about the noise currently affecting the system. Commonly, circle detection performed by the Circular Hough Transform [29], which employs edge information to infer locations and radius values. Peak detection is then performed by averaging, filtering and histo-gramming the transformed space. The overall approach requires a large storage space given the required 3-D cells to cover all parameters ($x$, $y$, $r$) and a high computational complexity which yields a low processing speed. The accuracy of the extracted parameters for the detected circle is poor, particularly under the presence of noise [3]. For a digital image holding a significant width and height and a densely populated edge pixel map, the required processing time for Circular Hough Transform makes it prohibitive to be deployed in real time applications.

In order to overcome this problem, some other researchers have proposed new approaches based on the Hough Transform including the probabilistic Hough Transform [40], the randomized Hough Transform (RHT) [46] and the Fuzzy Hough Transform [15]. Alternative other transformations have also been presented in literature as the one proposed by Becker in [6].

Recently, researchers have investigated evolutionary algorithms as an alternative way to perform circle detection. Such approaches involve the use of Genetic Algorithms (GA) [15], the Bacterial Foraging Algorithm Optimizer (BFAO) [16] and the Differential Evolution (DE) [17]. The advantage of these methods is that no assumptions are made about the objective function or the noise, yielding accurate solutions despite challenging and ambiguous environments. However, the improved accuracy often comes at a cost of an increased computational cost.

On the other hand, the Harmony Search Algorithm (HSA) is a new meta-heuristic method developed by Geem et al. [18] which has been inspired by musical performances from a musician seeking for a better state of harmony. In HSA, the solution vector is analogous to the harmony in music as local and global search schemes are analogous to musician's improvisations. In comparison to other meta-heuristics in the literature, HSA imposes fewer mathematical requirements as it can be easily adapted for solving several sorts of engineering optimization challenges [19,20]. Furthermore, numerical comparisons have demonstrated that the evolution for the HSA is faster than GA [19,21,22], capturing further attention as it has been successfully applied to solve a wide range of practical optimization problems, such as structural optimization, parameter estimation of the nonlinear Muskingum model, design optimization of water distribution networks, vehicle routing, combined heat and power economic dispatch, design of steel frames, bandwidth-delay-constrained least-cost multicast routing, transport energy modeling, among others [21–36].

This paper presents an algorithm for the automatic detection of circular shapes from complicated and noisy images with no consideration of the conventional Hough transform principles. The proposed algorithm is based on the novel HSA. The algorithm uses the encoding of three non-collinear edge points as candidate circles (Harmonies) in the edge-only image of the scene. An objective function evaluates (quality harmony) if such candidate circles are actually present in the edge image. Guided by the values of this objective function, the set of encoded candidate circles are evolved through the HSA so that they can fit into the actual circles within the edge map of the image (optimal harmony). The approach generates a sub-pixel circle detector which can effectively identify circles in real images despite circular objects exhibiting a significant occluded portion. Experimental evidence shows the effectiveness of such method for detecting circles under different conditions. Comparison to one state-of-the-art GA-based method [15], the BFAO detector [16] and the RHT algorithm [12] over multiple images demonstrates a better performance from the proposed method.

This paper is organized as follows: Section 2 provides information about the HSA while Section 3 depicts the implementation of the proposed circle detector. The study analyses the multi-circle detection in Section 4 with some conclusions being discussed in Section 5.

## 2. Harmony search algorithm

### 2.1. The Harmony Search Algorithm

In the basic HSA algorithm, each solution is called a ''harmony'' and is represented by an n-dimension real vector. An initial population of harmony vectors are randomly generated and stored within a harmony memory (HM). A new candidate harmony is thus generated from all solutions in the HM by using a memory consideration rule, a pitch adjustment rule and a random re-initialization. The HM is updated by comparing the new candidate harmony and the worst harmony vector in the HM. The worst harmony vector is replaced by the new candidate vector showing the better performance within the HM. The above process is repeated until a certain termination criterion is met. The basic HSA algorithm consists of three basic phases: initialization, improvisation of a harmony vector and updating the HM. The following discussion addresses details about each stage.

#### 2.1.1. Initializing the problem and algorithm parameters

In general, the global optimization problem can be summarized as follows: min $f(\mathbf{x})$ : $x(j) \in [l(j), u(j)], j = 1, 2, \ldots, n$, where $f(\mathbf{x})$ is the objective function, $\mathbf{x} = (x(1), x(2), \ldots, x(n))$ is the set of design variables, $n$ is the number of design variables, and $l(j)$ and $u(j)$ are the lower and upper bounds for the design variable $x(j)$, respectively. The parameters for HSA are the harmony memory size, i.e. the number of solution vectors lying on the harmony memory (*HMS*), the harmony-memory consideration rate (*HMCR*), the pitch adjusting rate (*PAR*), the distance bandwidth (*BW*) and the number of improvisations (*NI*) which represents the total number of function evaluations. It is obvious that a smart selection for HSA parameters would enhance the algorithm's ability to search for the global optimum under a high convergence rate.

#### 2.1.2. Initializing the harmony memory (HM)

The *HM* consists of *HMS* harmony vectors. Let $\mathbf{x}_i = \{x_i(1), x_i(2), \ldots, x_i(n)\}$ represent the *i*th randomly-generated harmony vector: $x_i(j) = l(j) + (u(j) - l(j)) \cdot r$ for $j = 1, 2, \ldots, n$ and $i = 1, 2, \ldots, HMS$, where $r$ is a uniform random number between 0 and 1. Then, the HM matrix is filled with the *HMS* harmony vectors as follows:

$$HM = \begin{bmatrix} \mathbf{x}_1 \\ \mathbf{x}_2 \\ \vdots \\ \mathbf{x}_{HMS} \end{bmatrix} \quad (1)$$

#### 2.1.3. Building a new harmony vector

A new Harmony vector $\mathbf{x}_{new}$ is built by applying three rules: the memory consideration rule, the pitch adjustment and the random selection. First of all, a uniform random number $r_1$ is generated within the range [0, 1]. If $r_1$ is less than *HMCR*, the decision variable $x_{new}(j)$ is generated through memory consideration; otherwise, $x_{new}(j)$ is obtained from a random selection, i.e. random re-initialization between the search bounds. For memory consideration, $x_{new}(j)$ is selected from any harmony vector $i$ in $\{1, 2, \ldots, HMS\}$. Second of all, each decision variable $x_{new}(j)$ will undergo a pitch adjustment under a probability of *PAR* if it is updated by memory consideration. The pitch adjustment rule is given as follows:

$$x_{new}(j) = x_{new}(j) \pm r \cdot BW \quad (2)$$

where *r* is a uniform random number between 0 and 1.

*2.1.4. Update harmony memory*

After a new harmony vector $\mathbf{x}_{new}$ is generated, the harmony memory is updated by the survival of the fit competition between $\mathbf{x}_{new}$ and the worst harmony vector $\mathbf{x}_w$ in the *HM*. Therefore $\mathbf{x}_{new}$ will replace $\mathbf{x}_w$ and become a new member of the *HM* in case the fitness value of $\mathbf{x}_{new}$ is better than the fitness value of $\mathbf{x}_w$.

*2.1.5. Computational procedure*

The computational procedure of the basic HAS can be summarized as follows [1]:

Step 1: Set the parameters *HMS, HMCR, PAR, BW* and *NI*.
Step 2: Initialize the *HM* and calculate the objective function value of each harmony vector.
Step 3: Improvise a new harmony $\mathbf{x}_{new}$ as follows:
  for (*j* = 1 to *n*) do
    if ( $r_1$ < *HMCR*) then
      $x_{new}(j) = x_a(j)$ where $a \in (1, 2, \ldots, HMS)$
      if ( $r_2$ < *PAR*) then
        $x_{new}(j) = x_{new}(j) \pm r_3 \cdot BW$ where $r_1, r_2, r_3 \in (0,1)$
      end if
    else
      $x_{new}(j) = l(j) + r \cdot (u(j) - l(j))$, where $r \in (0,1)$
    end if
  end for
Step 4: Update the *HM* as $\mathbf{x}_w = \mathbf{x}_{new}$ if $f(\mathbf{x}_{new}) < f(\mathbf{x}_w)$
Step 5: If *NI* is completed, return the best harmony vector $\mathbf{x}_b$ in the *HM*; otherwise go back to step 3.

## 3. Circle detection using HSA

At this work, circles are represented by a well-known second degree equation (see Equation 3) that passes through three points in the edge map. Pre-processing includes a classical Canny edge detector which uses a single-pixel contour marker and stores the location for each edge point. Such points are the only potential candidates to define circles by considering triplets. All the edge points are thus stored within a vector array $P = \{p_1, p_2, \ldots, p_{E_p}\}$ with $E_p$ being the total number of edge pixels in the image. The algorithm saves the $(x_i, y_i)$ coordinates for each edge pixel $p_i$ within the edge vector.

In order to construct each circle candidate, i.e. Harmonies within the HSA-framework, the indexes *i, j* and *k* of three non-collinear edge points must be combined, assuming that the circle's contour goes through points $p_i$, $p_j$, $p_k$. A number of candidate solutions are generated randomly for the initial set of the harmony memory (*HM*). Solutions will thus evolve through the application of the HSA upon harmonies until the optimal harmony (minimum) is reached, considering the best harmony of *HM* as the solution for the circle detection problem.

An overview of the required steps to formulate the circle detection task under the HSA optimization is presented below.

*3.1. Individual representation*

Each candidate solution **C** (harmony) uses three edge points. Under such representation, edge points are stored following a relative positional index within the edge array *P*. In turn, the procedure will encode a harmony as the circle that passes through three points $p_i$, $p_j$ and $p_k$ ($\mathbf{C} = \{p_i, p_j, p_k\}$). Each circle **C** is thus represented by three parameters $x_0$, $y_0$ and *r*, being $(x_0, y_0)$ the centre (*x*, *y*) coordinates for the circle and *r* its radius. The equation of the circle passing through the three edge points can thus be computed as follows:

$$(x - x_0)^2 + (y - y_0)^2 = r^2 \tag{3}$$

considering

$$\mathbf{A} = \begin{bmatrix} x_j^2 + y_j^2 - (x_i^2 + y_i^2) & 2 \cdot (y_j - y_i) \\ x_k^2 + y_k^2 - (x_i^2 + y_i^2) & 2 \cdot (y_k - y_i) \end{bmatrix} \quad \mathbf{B} = \begin{bmatrix} 2 \cdot (x_j - x_i) & x_j^2 + y_j^2 - (x_i^2 + y_i^2) \\ 2 \cdot (x_k - x_i) & x_k^2 + y_k^2 - (x_i^2 + y_i^2) \end{bmatrix}, \tag{4}$$

$$x_0 = \frac{\det(\mathbf{A})}{4((x_j - x_i)(y_k - y_i) - (x_k - x_i)(y_j - y_i))}, \quad y_0 = \frac{\det(\mathbf{B})}{4((x_j - x_i)(y_k - y_i) - (x_k - x_i)(y_j - y_i))}, \tag{5}$$

and

$$r = \sqrt{(x_0 - x_d)^2 + (y_0 - y_d)^2}, \tag{6}$$

being det(.) the determinant and $d \in \{i, j, k\}$. Figure 1 illustrates the parameters defined by Equations 3 to 6.

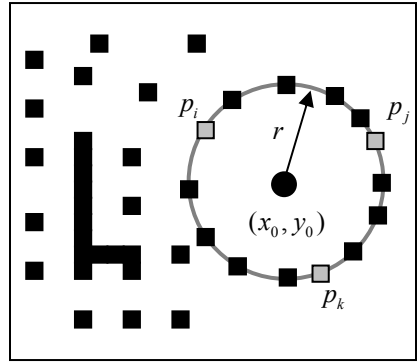

**Fig. 1.** Circle candidate (individual) built from the combination of points $p_i$, $p_j$ and $p_k$.

Therefore the shape parameters for the circle, [$x_0$, $y_0$, *r*] are represented as a transformation *T* with respect to edge vector indexes *i*, *j* and *k*.

$$[x_0, y_0, r] = T(i, j, k) \tag{7}$$

with *T* being the transformation calculated after the previous computations of $x_0$, $y_0$, and *r*. By considering each index as harmony parameter, it is feasible to apply the HAS seeking for appropriate circular parameters. The approach reduces the search space by eliminating unfeasible solutions.

*3.2 Objective function*

Optimization refers to the choosing of the best element from one set of available alternatives. In the simplest case, it means to minimize an objective function or error by systematically choosing variables values from valid ranges. In order to calculate the error produced by a candidate solution **C**, the circumference coordinates are calculated as a virtual shape which, in turn, must also be validated, i.e. if it really exists in the edge image. The test set is represented by $S = \{s_1, s_2, \ldots, s_{N_s}\}$, where $N_s$ are the number of points over which the existence of an edge point, corresponding to **C**, should be tested.

The set $S$ is generated by the midpoint circle algorithm [37]. The Midpoint Circle Algorithm (MCA) is a searching method that seeks for minimal required points for drawing a circle. Any point $(x, y)$ on the boundary of the circle with radius $r$ satisfies the equation $f_{Circle}(x,y) = x^2 + y^2 - r^2$. However, MCA avoids computing square-root calculations by comparing pixel separation distances. A method for direct distance comparison is to test the halfway position between two pixels (sub-pixel distance) to determine if this midpoint is inside or outside the circle boundary. If the point lies within the interior of the circle, the circle function is negative. Otherwise, if the point lies outside the circle, the circle function is positive. Therefore, the error involved in locating pixel positions using the midpoint test is limited to one-half of the pixel separation (sub-pixel precision). To summarize, the relative position of any point $(x, y)$ can be determined by checking the sign of the circle function:

$$f_{Circle}(x,y) \begin{cases} < 0 & \text{if } (x,y) \text{ is inside the circle boundary} \\ = 0 & \text{if } (x,y) \text{ is on the circle boundary} \\ > 0 & \text{if } (x,y) \text{ is outside the circle boundary} \end{cases} \quad (8)$$

The circle-function test in Eq. 8 is applied to mid-positions between pixels nearby the circle path at each sampling step. Figure 2a shows the midpoint between the two candidate pixels at sampling position $x_k$.

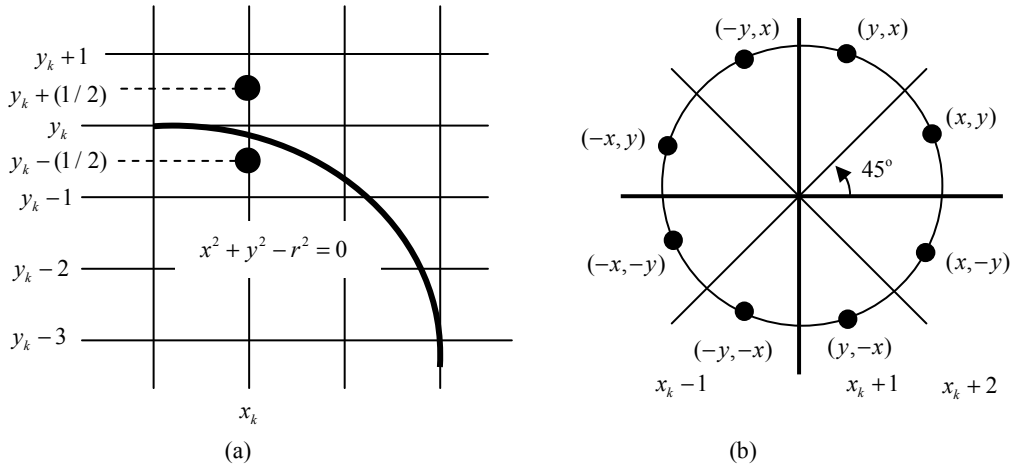

(a)      (b)

**Fig. 2.** Midpoint circle calculation: (a) Midpoint between candidate pixels at sampling position $x_k$ along a circular path. (b) Symmetry of a circle: calculation of a circle point $(x, y)$ for one octant yielding the circle points shown for other seven octants.

In MCA the computation time is reduced by considering the symmetry of circles. Circular sections in adjacent octants within one quadrant are symmetric with respect to the 45° line dividing the two octants. These symmetry conditions are illustrated in Figure 2b, where a point at position $(x, y)$ lying on a one-eighth circle sector is mapped into the seven circle points in the other octants of the ***xy*** plane. Taking advantage of the circle symmetry, it is possible to generate all pixel positions around a circle by calculating only the points within the sector from $x = 0$ to $x = y$. Thus, at this paper, the MCA is used to calculate the required $S$ points that represent the circle candidate **C**. The algorithm can be considered the quickest providing a sub-pixel precision [38]. However, in order to protect the MCA operation, it is important to assure that points lying outside the image plane must not be considered in *S*.

The objective function J(**C**) represents the matching error produced between the pixels *S* of the circle candidate **C** (harmony) and the pixels that actually exist in the edge image, yielding:

$$J(\mathbf{C}) = 1 - \frac{\sum_{v=1}^{Ns} E(x_v, y_v)}{Ns} \quad (9)$$

where $E(x_i, y_i)$ is a function that verifies the pixel existence in $(x_v, y_v)$, with $(x_v, y_v) \in S$ and $N_s$ being the number of pixels lying on the perimeter corresponding to **C** currently under testing. Hence, function $E(x_v, y_v)$ is defined as:

$$E(x_v, y_v) = \begin{cases} 1 & \text{if the pixel } (x_v, y_v) \text{ is an edge point} \\ 0 & \text{otherwise} \end{cases} \quad (10)$$

A value near to zero of *J*(**C**) implies a better response from the "circularity" operator. Figure 3 shows the procedure to evaluate a candidate solution **C** with its representation as a virtual shape *S*. In Figure 3b, the virtual shape is compared to the original image, point by point, in order to find coincidences between virtual and edge points. The virtual shape is built from points $p_i$, $p_j$ and $p_k$ shown by Fig. 3a, gathering 56 points (*Ns*= 56) with only 18 of such points existing in both images (shown as blue points plus red points in Fig. 3c) yielding: $\sum_{v=1}^{Ns} E(x_v, y_v) = 18$ and therefore $J(\mathbf{C}) \approx 0.67$.

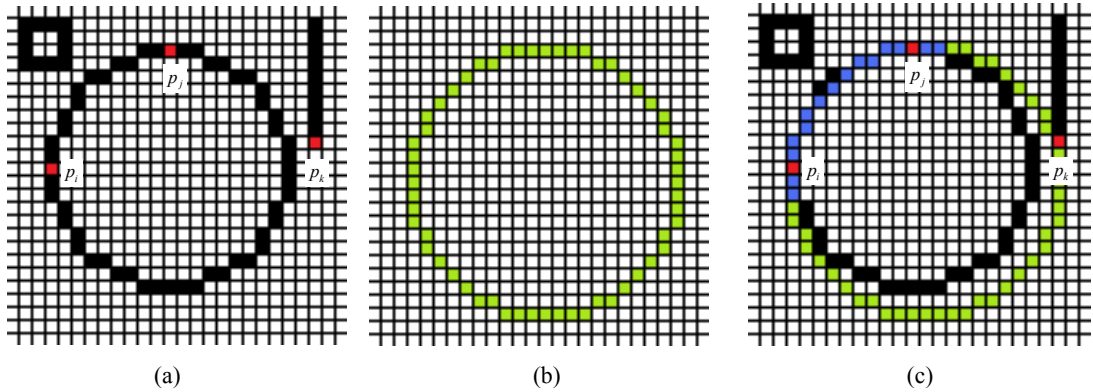

(a)  (b)  (c)

**Fig. 3.** Evaluation of candidate solutions **C**: the image in (a) shows the original image while (b) presents the virtual shape generated including points $p_i$, $p_j$ and $p_k$. The image in (c) shows coincidences between both images marked by blue or red pixels while the virtual shape is also depicted in green.

### 3.3. Implementation of HSA for circle detection

The implementation of HSA can be summarized into the following steps:

| Step 1: | Setting the HSA parameters. Initializing the harmony memory with *HMS* individuals where each decision variable $p_i$, $p_j$ and $p_k$ of the candidate circle $\mathbf{C}_a$ is set randomly |
|---|---|
| | within the interval $[1, E_p]$. All values must be integers. Considering $a \in (1,2,\ldots, HMS)$. |
| **Step 2:** | Evaluating the objective value $J(\mathbf{C}_a)$ for all *HMS* individuals, and determining the $\mathbf{C}_w$ showing the worst objective value. |
| **Step 3:** | Improvising a new harmony $\mathbf{C}_{new}$ such that: <br> for ($j$ = 1 to 3) do <br>   if ($r_1 <$ *HMCR*) then <br>     $C_{new}(j) = C_a(j)$ where $a \in (1,2,\ldots, HMS)$ <br>     if ($r_2 <$ *PAR*) then <br>       $C_{new}(j) = C_{new}(j) \pm r_3 \cdot BW$ where $r_1, r_2, r_3 \in (0,1)$ <br>     end if <br>   else <br>     $C_{new}(j) = 1 + \text{round}(r \cdot E_p)$, where $r \in (0,1)$ <br>   end if <br> end for |
| **Step 4:** | Update the *HM* as $\mathbf{C}_w = \mathbf{C}_{new}$ if $J(\mathbf{C}_{new}) < J(\mathbf{C}_w)$ |
| **Step 5:** | If *NI* is completed then return the best harmony vector $\mathbf{C}_b$ in the *HM* (a circle contained in the image); otherwise go back to step 2. |

Figure 4 shows the outcome of a given detection test. The input image (Fig. 4a) has a resolution of 256 x 256 pixels and shows a hand-drawn circle. Figure 4b presents the detected circle through a blue overlay.

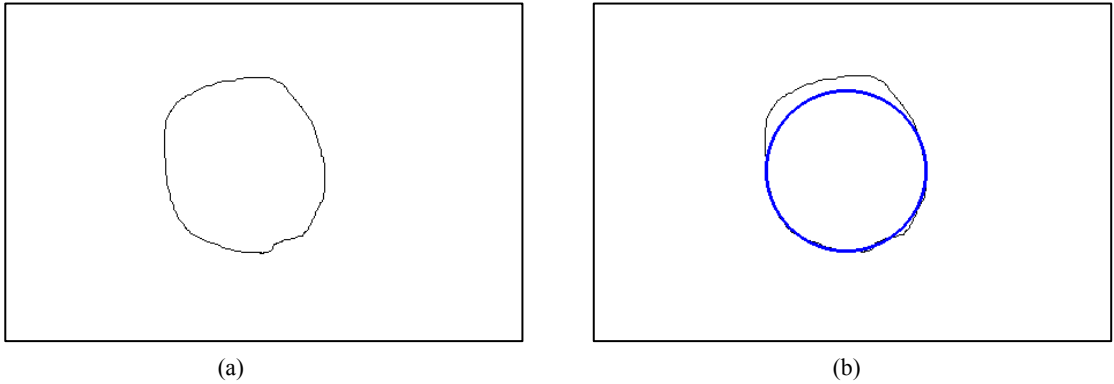

(a)           (b)

**Fig. 4.** HSA processing test: image in (a) shows a hand-drawn circular shape while (b) shows the approximated circular shape through a blue overlay.

## 4. Experimental results

In order to evaluate the performance of the circle detector proposed in this paper, several experimental tests have been developed. Table 1 presents the parameters of HSA used at this work which have been experimentally determined and kept for all test images through all experiments.

| HMS | HMCR | PAR | BW | NI |
|---|---|---|---|---|
| 100 | 0.7 | 0.3 | 2 | 200 |

**Table 1.** Parameter setup for the HSA detector

All the experiments are performed on a Pentium IV 2.5 GHz computer under C language programming. All the images are preprocessed by the standard Canny edge-detector using the image-processing toolbox for MATLAB R2008a.

*4.1 Circle localization*

The experimental setup includes the use of 60 synthetic and natural images. In the synthetic case, the images have been generated drawing only a randomly located circle. The parameters under detection are the center of the circle (*x*, *y*) and its radius (*r*). The algorithm is set to 200 iterations for each test image. In all cases the algorithm is able to detect the required parameters. The detection is shown to be robust to translation and scale conserving a reasonably low elapsed time (typically under 1ms). Figure 5b shows the results of the circle detection for a synthetic image.

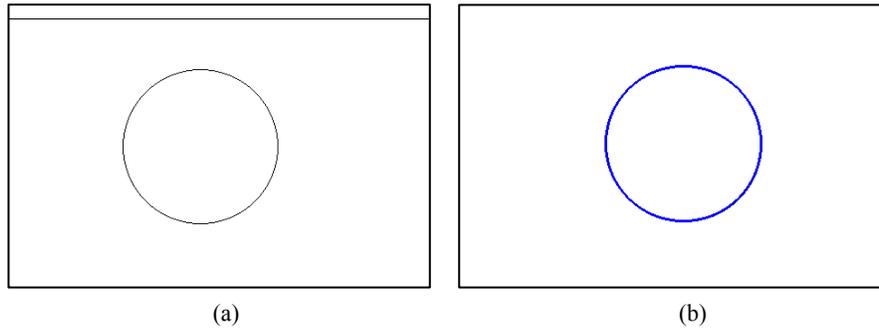

(a)          (b)
**Fig. 5.** Circle detection in synthetic images: (a) original circle image, (b) its corresponding detected circle.

On the other hand, natural-life images rarely contain perfect circles forcing the detection algorithm to approximate the circle that better adapts to the imperfect circle within the image. Such circle corresponds to the smallest error obtained from the objective function *J(*C*)*. Detection results have been statistically analyzed for comparison purposes. For instance, the detection algorithm is executed 50 times on the same image (Figure 6), yielding the same parameters $x_0 = 210$, $y_0 = 325$, and $r = 65$, which indicates that the proposed HSA is able to converge to the minimum solution. The experiment has considered 200 cycles.

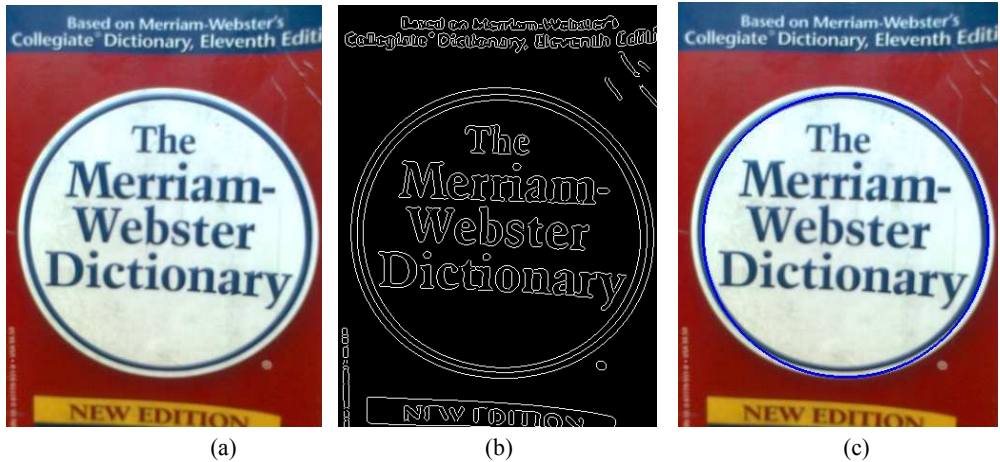

(a)          (b)          (c)

**Fig. 6.** A experiment on a real-life image: (a) shows the test image and (b) its corresponding edge map. The detected circle is shown in (c) through an overlay.

The circle detection algorithm described in this paper is also useful to approximate circular shapes from arc segments, occluded circular shapes or imperfect circles. This functionality is quite relevant considering that such shapes are common to typical computer vision problems. The proposed algorithm is able to find circle parameters that better approach the arc, occluded or imperfect circles. Figure 7 shows one example of this functionality considering a hand-drawn arc. Recalling that, at this paper, the detection process is approached as an optimization problem and that the objective function *J(C)* gathers the **C** points actually contained in the image, a smaller value of *J(C)* commonly refers to a circle while a greater value accounts for either an arc, an occluded circle or an imperfect circle. Such fact does not represent any trouble as circles would be detected first while other shapes would follow. In general, the detection of all kinds of circular shapes would only differ according to smaller or greater values of *J(C)*.

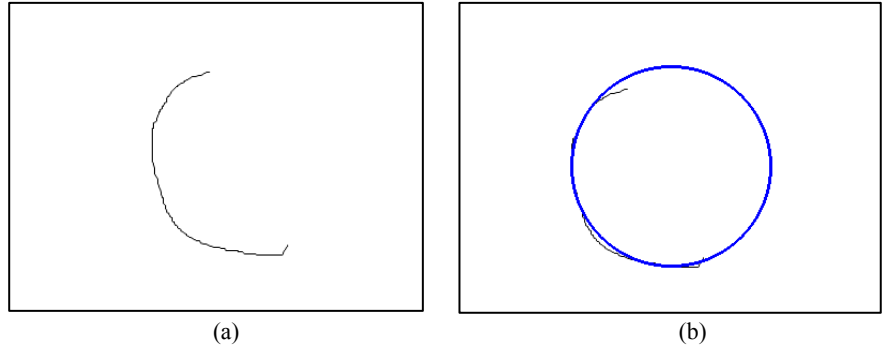

(a)          (b)

**Fig. 7.** HSA approximating circular arc sections: image in (a) shows the hand-drawn arc and (b) the detected circle.

*4.2. Multiple circle detection*

The HSA circle detector is also capable of detecting several circles embedded into a single image. The procedure is applied in the same way until the first circle is detected. It represents the circle with the minimum objective function value *J*(**C**). Therefore, such shape is masked (i.e. eliminated) on the primary edge-only image as the HSA circle detector operates again over the modified image. The procedure is repeated until the *J*(C) value reaches a minimum predefined threshold $M_{th}$ (typically 0.1). Finally, a subsequent validation of all detected circles follows by analyzing continuity of the detected circumference segments as proposed in [39]. If none of the detected shapes satisfies the $M_{th}$ criterion, the system simply reply: "no circle detected". Figure 8a shows one natural image containing several circles. For this case, the algorithm searched for the best circular shapes (greater than $M_{th}$). The edge-only image after the Canny algorithm application is also shown by Figure 8b, with Figure 6(c) presenting the detected circles. Likewise, Figure 9 shows the multi-circle detection performance of HSA, considering complex synthetic images. Figure 9(a) has been hand-drawn generated. The same image has been contaminated by adding noise as to increase the complexity in the detection process (see Figure 9c). Figures 9b and 9d show the detected circle after the application of HSA.

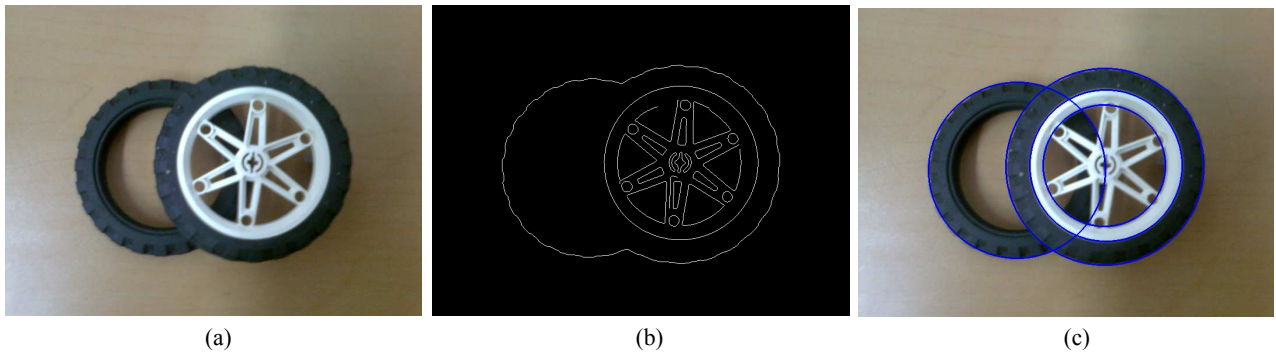

(a)          (b)          (c)

**Fig. 8.** Multiple circle detection over natural images: (a) the original image, (b) the edge image after applying the Canny algorithm and (c) the image portraying best detected circular shapes.

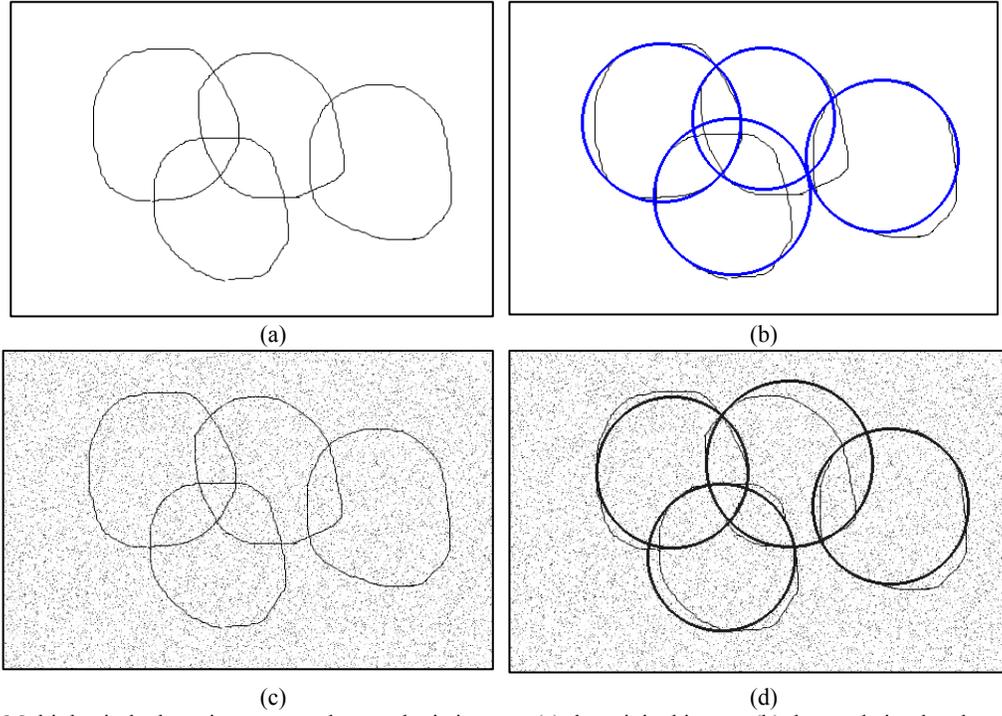

(a)　　　　　　　　　　　　　　　　　　(b)

(c)　　　　　　　　　　　　　　　　　　(d)

**Fig. 9.** Multiple circle detection on complex synthetic images: (a) the original image, (b) detected circular shapes, (c) the original noisy image and (d) detected circular shapes.

*4.3. Performance comparison*

In order to enhance the algorithm analysis, the HSA is compared to the BFAO and the GA circle detectors over an image set.

The GA algorithm follows the proposal of Ayala-Ramirez et al., which considers the population size as 70, the crossover probability as 0.55, the mutation probability as 0.10 and the number of elite individuals as 2. The roulette wheel selection and the 1-point crossover operator are both applied. The parameter setup and the fitness function follow the configuration suggested in [15]. The BFAO algorithm follows the implementation from [16] considering the experimental parameters as: $S=50$, $N_c = 100$, $N_s = 4$, $N_{ed} = 1$, $P_{ed} = 0.25$, $d_{attract} = 0.1$, $w_{attract} = 0.2$, $w_{repellant} = 10$, $h_{repellant} = 0.1$, $\lambda = 400$ and $\psi = 6$. Such values are found to be the best configuration set according to [16].

Images rarely contain perfectly-shaped circles. Therefore, with the purpose of testing accuracy for a single-circle, the detection is challenged by a ground-truth circle which is determined from the original edge map. The parameters $(x_{true}, y_{true}, r_{true})$ representing the testing circle are computed using the Equations 3 to 6 for three circumference points over the manually-drawn circle. Considering the centre and the radius of the detected circle are defined as $(x_D, y_D)$ and $r_D$, the Error Score (***Es***) can be accordingly calculated as:

$$Es = \eta \cdot (|x_{true} - x_D| + |y_{true} - y_D|) + \mu \cdot |r_{true} - r_D| \qquad (11)$$

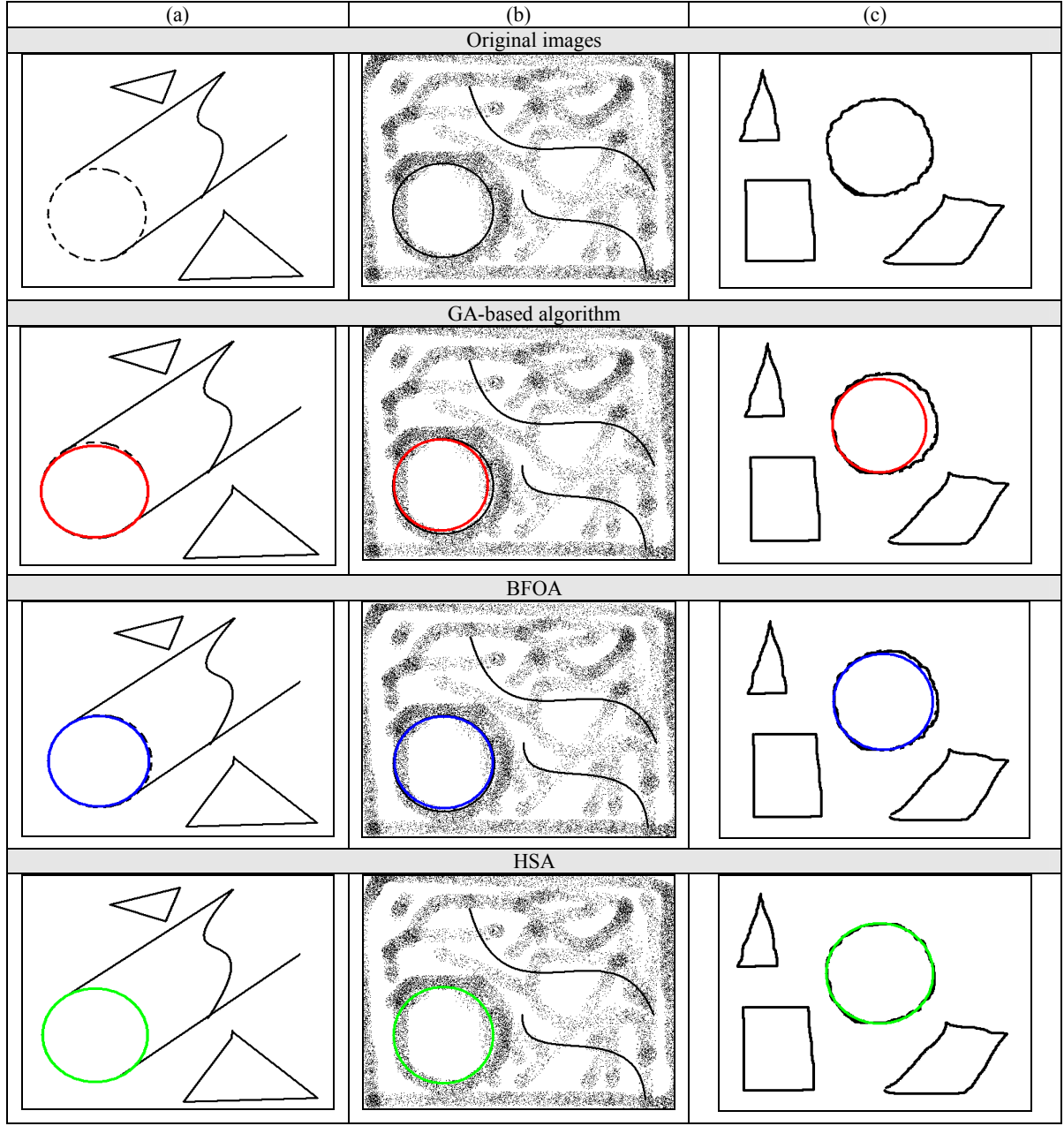

**Fig. 10.** Synthetic images and their detected circles after applying the GA-based algorithm, the BFOA method and the proposed HSA.

The central point difference $(|x_{true} - x_D| + |y_{true} - y_D|)$ represents the centre shift for the detected circle as it is compared to a benchmark circle. The radio mismatch $(|r_{true} - r_D|)$ accounts for the difference between their radii. $\eta$ and $\mu$ represent two weighting parameters which are to be applied separately to the central point difference and to the radio mismatch for the final error $Es$. At this work, they have been chosen as $\eta = 0.05$ and $\mu = 0.1$ aiming to ensure that the radii difference would be strongly weighted in comparison to the difference of central circular positions between the manually detected and the machine-detected circles. In case the value of $Es$ is less than 1, the algorithm gets a success; otherwise it has failed on detecting the edge-circle. Notice that for $\eta = 0.05$ and $\mu = 0.1$, it yields $Es<1$ which accounts for a maximal tolerated difference on radius length of 10 pixels, whereas the maximum mismatch for the centre location can be up to 20 pixels. In general, the success rate (SR) can thus be defined as the percentage of reaching success after a certain number of trials.

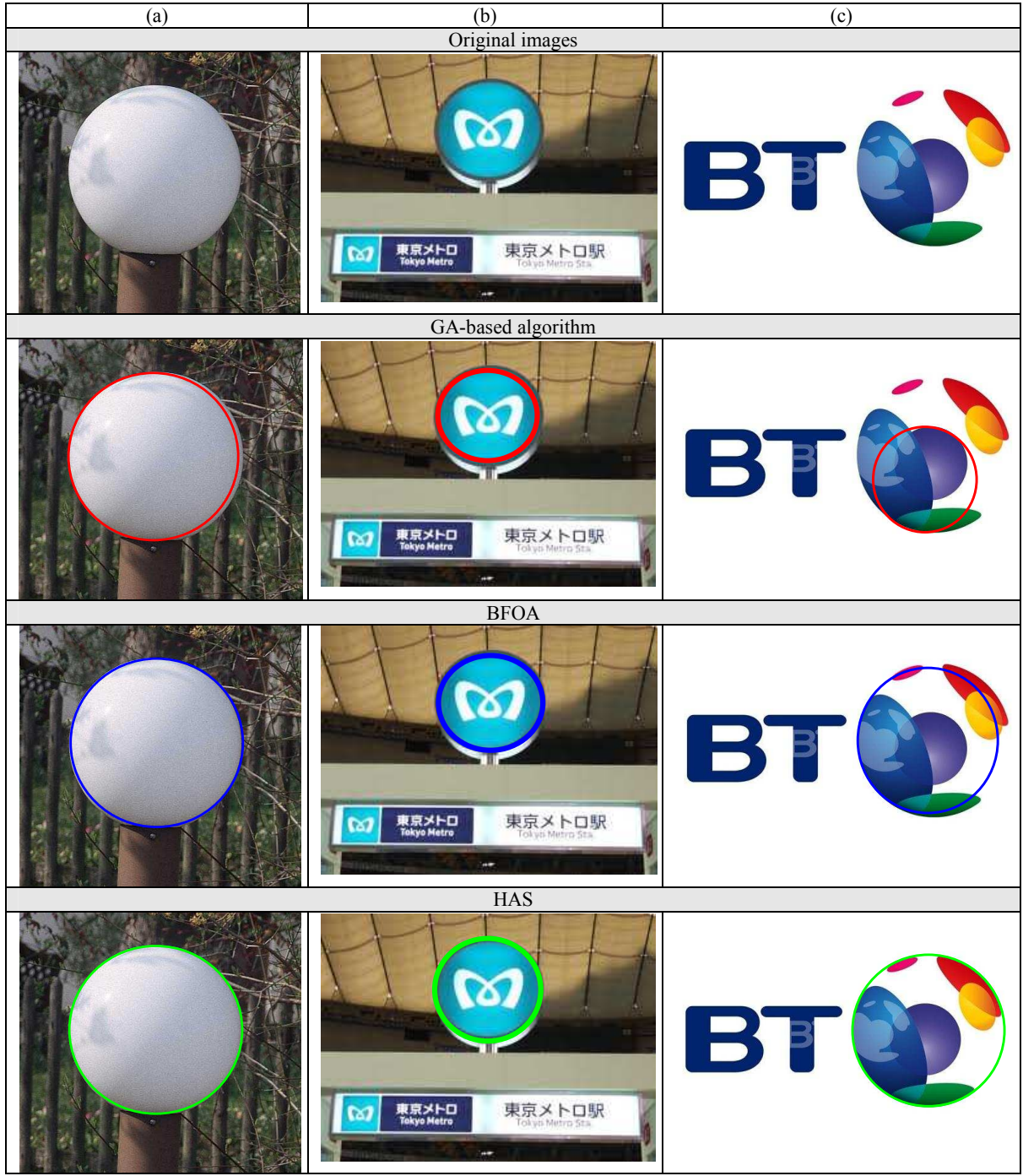

**Fig. 11.** Real-life images and their detected circles for: GA-based algorithm, the BFOA method and the proposed HAS.

Figure 10 shows three synthetic images and the resulting images after applying the GA-based algorithm [15], the BFOA method [16] and the proposed approach. Figure 11 presents experimental results considering three natural images. The performance is analyzed by considering 35 different executions for each algorithm. Table 2 shows the averaged execution time, the success rate in percentage and the averaged Error Score ($Es$), considering six test images (shown by Figures 10 and 11). The best entries are bold-cased in Table 2. Close inspection reveals that the proposed method is able to achieve the highest success rate keeping the smallest error yet requiring less computational time for the most cases.

A non-parametric statistical significance-proof called Wilcoxon's rank sum test for independent samples [40-42] has been conducted at the 5% significance level on the Error Score ($Es$) data of Table 2. Table 3 reports the $p$-values produced by Wilcoxon's test for the pair-wise comparison between the Error Score ($Es$) from two groups. One group corresponds to HSA vs. GA and the other corresponds to an HSA vs. BFOA, one at a time. As a null hypothesis, it is assumed that there is no significant difference between mean values of such two groups. The alternative hypothesis considers a significant difference between mean values of both groups. All $p$-values reported in the table are less than 0.05 (5% significance level) which is a strong evidence against the null hypothesis, indicating that the best HSA mean values for the performance are statistically significant.

| Image | Averaged execution time ± Standard deviation (s) | | | Success rate (SR) (%) | | | Averaged Es ± Standard deviation | | |
|---|---|---|---|---|---|---|---|---|---|
| | GA | BFOA | HSA | GA | BFOA | HSA | GA | BFOA | HSA |
| Synthetic images | | | | | | | | | |
| (a) | 2.51±(0.31) | 1.12±(0.45) | **0.31±(0.12)** | 98 | **100** | 100 | 0.45±(0.022) | 0.30±(0.033) | **0.20±(0.021)** |
| (b) | 3.56±(0.44) | 3.02±(0.32) | **0.33±(0.25)** | 98 | 98 | **100** | 0.61±(0.022) | 0.41±(0.034) | **0.19±(0.035)** |
| (c) | 4.67±(0.34) | 3.92±(0.21) | **0.30±(0.21)** | 75 | 90 | **100** | 0.56±(0.029) | 0.46±(0.051) | **0.21±(0.012)** |
| Natural Images | | | | | | | | | |
| (a) | 5.44±(0.21) | 4.23±(0.34) | **0.52±(0.36)** | 98 | **100** | 100 | 0.40±(0.072) | 0.29±(0.041) | **0.20±(0.025)** |
| (b) | 6.11±(0.27) | 5.07±(0.14) | **0.51±(0.34)** | 80 | 96 | **100** | 0.64±(0.025) | 0.58±(0.037) | **0.27±(0.024)** |
| (c) | 7.21±(0.36) | 6.12±(0.31) | **0.54±(0.25)** | 80 | 88 | **97** | 0.88±(0.043) | 0.73±(0.037) | **0.37±(0.012)** |

**Table 2.** The averaged execution-time, success rate and the averaged error score for the GA-based algorithm, the BFOA method and the proposed HSA algorithm, considering six test images (shown by Figures 8 and 9).

| Image | $p$-Value | |
|---|---|---|
| | HSA vs. GA | HSA vs. BFOA |
| Synthetic images | | |
| (a) | 1.9456e-004 | 1.9234e-004 |
| (b) | 1.6341e-004 | 1.8892e-004 |
| (c) | 1.4562e-004 | 1.8648e-004 |
| Natural Images | | |
| (a) | 1.8872e-004 | 1.8945e-004 |
| (b) | 1.5124e-004 | 1.6725e-004 |
| (c) | 1.3289e-004 | 1.4289e-004 |

**Table 3.** $p$-values produced by Wilcoxon's test comparing HSA to GA and BFOA over an averaged $Es$ from Table 2.

Figure 12 demonstrates the relative performance of HSA in comparison to the RHT algorithm as it is described in [12]. Images from the test are complicated as they contain different noise conditions. Table 4 reports the corresponding averaged execution time, the success rate (in %), and the averaged error score (calculated following Eq. 11) for HSA and RHT algorithms over three test images shown by Figure 12. Table 4 also shows the performance loss as noise conditions vary. Yet the HSA algorithm holds its performance under the same circumstances.

| Image | Average time ± Standard deviation (s) | | Success rate (SR) (%) | | Average Es ± Standard deviation | |
|---|---|---|---|---|---|---|
| | RHT | HSA | RHT | HSA | RHT | HSA |
| (I) | 4.32±(0.51) | **0.30±(0.34)** | 100 | 100 | 0.18±(0.030) | **0.21±(0.026)** |
| (II) | 7.34±(0.21) | **0.33±(0.25)** | 85 | **100** | 0.76±(0.037) | **0.24±(0.031)** |
| (III) | 9.61±(0.33) | **0.32±(0.21)** | 50 | **100** | 1.62±(0.044) | **0.23±(0.043)** |

**Table 4.** Average time, success rate and averaged error score for the HSA and the RHT, considering three test images

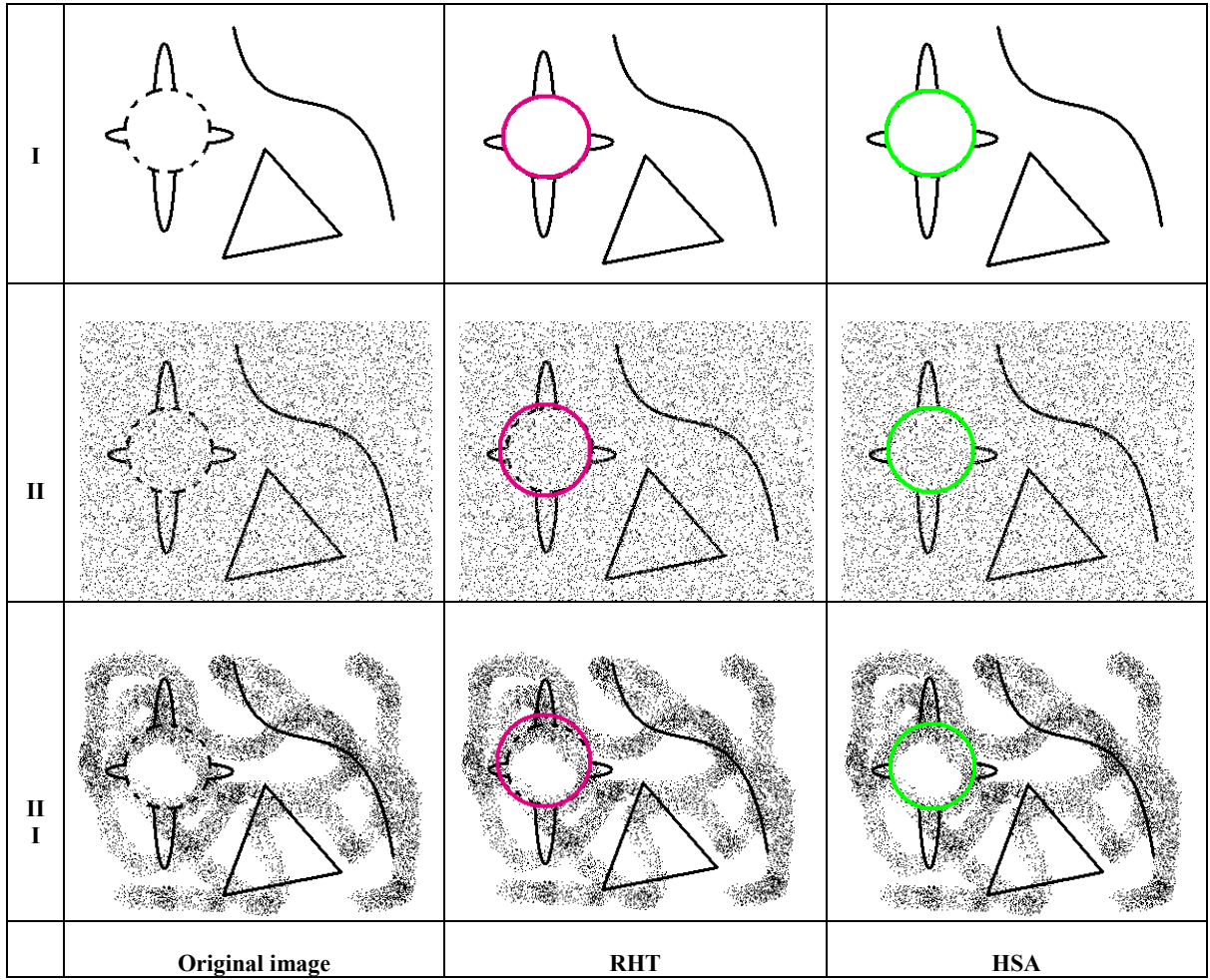

**Fig. 12.** Comparative performance of the RHT and the HSA

## 5. Conclusions

This work has presented an algorithm for the automatic detection of circular shapes from complicated and noisy images with no consideration of the conventional Hough Transform principles. The proposed method is based on the novel Harmony Search Algorithm (HSA). To the best of our knowledge, the HSA has not been applied to any image processing task until date. The algorithm uses the encoding of three non-collinear edge points as circle candidates (harmonies) that have been taken from the edge-only image of the scene. An objective function (harmony quality) evaluates if a given circle candidate is actually present in the edge image. Guided by the values of the objective function, the set of encoded candidate circles is evolved using the HSA so that they can fit into the actual circles in the edge map of the image (optimal harmony). As it can be observed from results shown in Figures 10-12, our approach detects a circle embedded into complex images with little visual distortion despite the presence of noisy background pixels.

The paper approaches the circle detection as an optimization problem. Such view enables the algorithm to detect arcs, occluded circles and also matching imperfect circles. The HSA is capable of finding circle parameters according to $J(\mathbf{C})$, clearly opposing to other methods which make a review of all circle candidates for detecting occluded or imperfect circles.

In order to test the circle detection accuracy, a score function has been used (Eq. 11). It can objectively evaluate the mismatch between a manually detected circle and a machine-detected circle. We have demonstrated that the HSA method outperforms the GA (as described in [15]), the BFAO (as described in [16]) and the RHT (as described in [12]) within a statistically significant framework.

Classical Hough Transform methods for circle detection use three edge points to cast a vote for potential circular shapes within the parameter space. However, they would require a huge amount of memory and longer computational times to obtain a sub-pixel resolution. Moreover, HT-based methods rarely find a precise parameter set for a circle in the image [43]. In our approach, the detected circles are directly obtained from Equations 6 to 9 still reaching sub-pixel accuracy.

Although Figures 8 and 9 indicate that the HSA method can yield better results on complicated and noisy images in comparison to the GA, BFAO and the RHT methods, the aim of our paper is not intended to beat all the circle detector methods which have been proposed earlier, but to show that the Harmony Search Systems can effectively serve as an attractive evolutionary alternative to successfully extract circular shapes from images.